\pdfoutput=1
\documentclass[conference, compsoc]{IEEEtran}
\usepackage{graphicx,amssymb,amsmath}
\usepackage{balance}
\balance

\newtheorem{theorem}{Theorem}

\newtheorem{definition}[theorem]{Definition}
\newtheorem{algorithm}[theorem]{Algorithm}

\begin{document}

% Paper title: keep the \ \\ \LARGE\bf in it to leave enough margin.
\title{Heuristic Reasoning on Graph and Game Complexity of Sudoku}

\author{Zhe Chen\\
LATTIS, INSA, University of Toulouse\\
135 Avenue de Rangueil, 31077 Toulouse, France\\
Email: zchen@insa-toulouse.fr}

\date{}
\maketitle

\begin{abstract}
The Sudoku puzzle has achieved worldwide popularity recently, and
attracted great attention of the computational intelligence
community. Sudoku is always considered as Satisfiability Problem or
Constraint Satisfaction Problem. In this paper, we propose to focus
on the essential graph structure underlying the Sudoku puzzle.
First, we formalize Sudoku as a graph. Then a solving algorithm
based on heuristic reasoning on the graph is proposed. The related
r-Reduction theorem, inference theorem and their properties are
proved, providing the formal basis for developments of Sudoku
solving systems. In order to evaluate the difficulty levels of
puzzles, a quantitative measurement of the complexity level of
Sudoku puzzles based on the graph structure and information theory
is proposed. Experimental results show that all the puzzles can be
solved fast using the proposed heuristic reasoning, and that the
proposed game complexity metrics can discriminate difficulty levels
of puzzles perfectly.

%\keywords{Sudoku, knowledge representation, heuristic reasoning,
%graph, game complexity}
\end{abstract}

\section{Introduction}
The Sudoku puzzle has become popular in Japan since 1986, and became
world-wide popular in 2005. The completion rules of the puzzle are
rather simple, however the required reasoning techniques for solving
the puzzle are often complex. Because Sudoku puzzle is a good
example of various reasoning approaches in Artificial Intelligence
area, it has attracted great attention of the computational
intelligence community
\cite{Caine2006}\cite{Henz}\cite{Lynce}\cite{Reeson}\cite{Simonis}\cite{Weber}.

\begin{figure*}[tb]
  % Requires \usepackage{graphicx}
  \centering
  \includegraphics[bb=0 0 255 255, scale=.5]{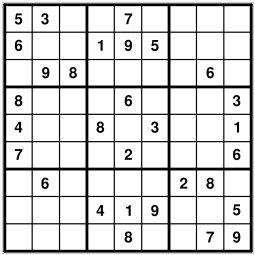}
  \includegraphics[bb=0 0 254 254, scale=.5]{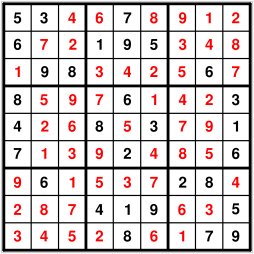}\\
  \caption{A Sudoku Puzzle}\label{Fig:Sudoku}
\end{figure*}

Fig. \ref{Fig:Sudoku} is an example of the Sudoku puzzle. The left
part provides a Sudoku puzzle, and its solution is shown on the
right. We define the notations as follows:

\begin{definition}
The \textbf{Sudoku puzzle} is a $9 \times 9$ \textbf{grid}, which
comprises nine $3 \times 3$ \textbf{sub-grid}s, which comprises $3
\times 3$ \textbf{cell}s. Some cells are filled with numbers from 1
to 9, whereas others are left blank.
\end{definition}

\begin{definition}
The Sudoku puzzle is \textbf{solved} by filling all the blank cells
with numbers from 1 to 9, such that every row, every column and
every sub-grid contains each of the nine possible numbers. Sudoku
puzzle has exactly one solution.
\end{definition}

There are mainly three directions in the Sudoku puzzle studies:

1. Sudoku generation. One question is how to generate randomly a
Sudoku puzzle which has exactly one solution. Another question is
the estimation of the total number of the possible Sudoku puzzles
\cite{Felgenhauer2005}\cite{Felgenhauer2006}\cite{Russell}.

2. Sudoku solving. Up to now, most of the studies are on this
direction. People proposed several practical reasoning algorithms
based on different knowledge representations. \cite{Yato} shows that
the generalized Sudoku puzzle is NP-Complete.
\cite{Henz}\cite{Lynce}\cite{Weber} encode Sudoku as a
Satisfiability Problem (SAT), and use a general SAT solver to get
the solution. \cite{Reeson}\cite{Simonis} represents the Sudoku
puzzle as a Constraint Satisfaction Problem (CSP), and compares
different propagation schemes for solving Sudoku. However,
\cite{Henz} reported that some hard puzzles cannot be solved using
these methods, such as \cite{Inkala}\cite{Ravel}. In this paper, we
will formalize Sudoku as a graph, and develop a solving algorithm
based on heuristic reasoning on the graph.

3. Sudoku complexity evaluation. Most of the Sudoku designers
classify the difficulty levels of Sudoku puzzles as, for instance,
``easy", ``intermediate" or ``hard" (or levels 1, 2, 3) according to
their experiences. However, up to now there are few publications on
this subject. \cite{Lynce} only mentioned that the difficulty level
depends on the number of initial non-blank cells provided. This is a
very rough evaluation, and we will propose a more accurate
quantitative metric for difficulty evaluation based on the
complexity of the underlying graph structure.

This paper deals with the second and third aspects, in order to
provide a formal basis for the studies and developments of Sudoku
solving systems and difficulty evaluation. In section II, the graph
structure underlying Sudoku puzzle is defined. In section III,
heuristic reasoning on the graph is proposed, and its related
theorems, properties and algorithm are explained in detail. Section
IV focuses on how to evaluate the game complexity of the Sudoku
puzzle based on information theory. To assess our theory,
experiments are performed and the results are discussed in section
V. Conclusion and future works are presented at the end.

\section{The Sudoku Puzzle and Graph}

\begin{figure}[tb]
  % Requires \usepackage{graphicx}
  \centering
  \includegraphics[bb=0 0 470 481, scale=.5]{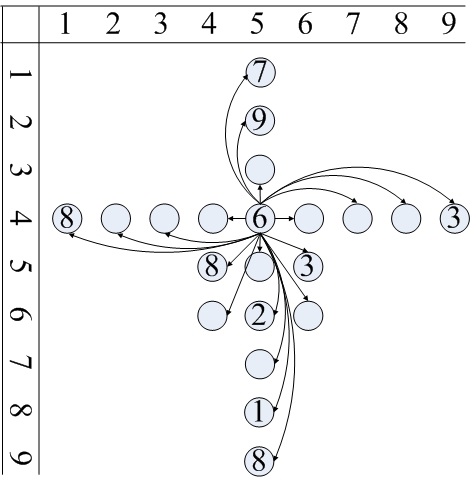}\\
  \caption{Part of the Sudoku Graph}\label{Fig:Sudoku_Graph}
\end{figure}

The Sudoku puzzle is essentially a directed graph. The cells are
vertices, and the constraints are directed edges. Fig.
\ref{Fig:Sudoku_Graph} shows part of the Sudoku graph related to
$cell_{45}$ of the puzzle in Fig. \ref{Fig:Sudoku}. The cells in the
same row, the same column and the same sub-grid as $cell_{45}$ are
vertices in the subgraph, and the edges represent the constrain
relationships between $cell_{45}$ and the connected cells. Notice
that the full graph is constructed by $9 \times 9$ such subgraphs
that are connected to be the entire graph. Initially, $cell_{45}$ is
filled with number 6, so we say that the set of possible values of
the vertex is $v_{45}=\{6\}$, and the filled value is $c_{45}=6$.

In Fig. \ref{Fig:Sudoku}, the vertex $v_{13}$ may be filled with 1,
2, or 4, so the candidates set is $v_{13}=\{1,2,4\}$, and the size
of the set is $|v_{13}|=3$. In the reasoning process, if we find
that $cell_{31}$ should be filled with 1, then we can remove 1 from
the candidates set $v_{13}$ (they are connected by an edge
$e_{31,13}$), so $v_{13}=\{2,4\}$. That is, the information of
possible candidates in each cell is propagated through the edges on
the graph, and change the values of candidates sets of connected
vertices. The graph underlying the Sudoku puzzle is formally defined
as follows.

\begin{definition}
The \textbf{Sudoku alphabet} is $\sum = \{1,2,3,4,5,6,7,8,9\}$.
\end{definition}

\begin{definition}
The \textbf{vertex candidates set} is $v_{ij}=\{x|x \in \sum$, and
$cell_{ij}$ is possible to be filled with the number $x$ with the
information we have$\}$.
\end{definition}

\begin{definition}\label{Def:filled}
If $|v_{ij}|=1$, the vertex $cell_{ij}$ is \textbf{filled} with the
number $c_{ij} \in v_{ij}$.
\end{definition}

\begin{definition}\label{Def:Sudoku_Graph}
The \textbf{Sudoku graph} is $G=(V,E)$, where the vertices set $V=\{
cell_{ij}|cell_{ij}$ is the cell in $i$-th row and $j$-th column, $1
\leq i,j \leq 9 \}$, the set of edges $E=\{ (cell_{ij}, cell_{kl}),
(cell_{kl}, cell_{ij})|$ there is the constraint $c_{ij} \neq
c_{kl}$ between $cell_{ij}$ and $cell_{kl}\}$. For simplification,
edge $(cell_{ij}, cell_{kl})$ is noted as $e_{ij,kl}$.
\end{definition}

\begin{definition}
\textbf{Initiation} of Sudoku Graph: $\forall cell_{ij}$, if
$cell_{ij}$ is filled with number $k$ in the puzzle as initial, then
$c_{ij}=k$, $v_{ij}=\{k\}$.
\end{definition}

Both the solving algorithm and game complexity proposed in this
paper are based on the graph structure. Ideally, player can solve
the puzzle in this way: player starts from any one of the vertices,
traverses all the vertices once and only once, along with the
directed edges. Each time when the player arrives at a vertex
$cell_{ij}$, he or she want to choose the correct number $x \in
v_{ij}$ to fill the cell $c_{ij}=x$, and output the number. The
player continues to traverse the remaining vertices by choosing one
of the edges, until all the vertices are visited. At the end, the
output of numbers is the solution.

However, how to ``choose the correct number $x \in v_{ij}$'' is the
essential uncertain problem. In the next section, a heuristic
reasoning approach is proposed to achieve this goal. And the
difficulty level of the Sudoku puzzle is evaluated based on the
concept of ``uncertainty of choice'' at each vertex of the graph,
since it is related to the magnitude of uncertainty in the
reasoning.

\section{Heuristic Reasoning on the Graph}
In order to present the heuristic reasoning on the Sudoku graph, we
define the concept of ``section'' of a Sudoku graph.

\begin{definition}
\textbf{Section $S$} is a complete subgraph in the Sudoku graph $G$.
\end{definition}

``Complete subgraph'' is a basic concept from discrete mathematics.
The definition means that a section is a subgraph of graph $G$, such
that every two vertices in a section are directly connected by two
directed edges of opposite directions. The vertices in a row, a
column or a sub-grid construct totally 27 ``sections''. Moreover,
they are the only sections in a Sudoku graph. Each vertex belongs to
3 sections.

For heuristic reasoning, the most obvious fact is that if a cell is
filled with a number, then the cells in the same section cannot be
filled with the same number. So this number can be removed from the
candidates sets. Because we only remove one element from the sets,
the process is called 1-Reduction.

\begin{theorem}[1-Reduction] $\forall i,j,k,l$, if $\exists |v_{ij}|=1$,
and there is an edge $e_{ij,kl}$, then $c_{kl} \in v_{kl}-v_{ij}$.
\end{theorem}

{\em proof:} According to Definition \ref{Def:filled},
\begin{equation}
c_{kl} \in v_{kl}
\end{equation}

There is an edge $e_{ij,kl}$, according to Definition
\ref{Def:Sudoku_Graph}, $c_{ij} \neq c_{kl}$. So,
\begin{equation}
c_{kl} \in v_{kl} - \{c_{ij}\}
\end{equation}

$|v_{ij}|=1$, so $v_{ij}=\{ c_{ij} \} $, we conclude $ c_{kl} \in
v_{kl}-v_{ij}$.\hfill $\Box$

Using this theorem, once a cell is filled with number $c_{ij}$, then
we can remove the number from the candidates sets of the connected
vertices.

However, sometimes we are not able to find such a vertex having
$|v_{ij}|=1$, but we are able to find $r$ vertices that share $r$
possible numbers. In this case, we can remove the $r$ numbers from
the candidates sets of the other vertices that are in the same
section as the $r$ vertices. Notice that $r$ numbers are reduced
from the sets. To formalize the information propagation on the
graph, we have a more general reasoning theorem:

\begin{theorem}[r-Reduction]\label{The:rR} In any section $S_k$, if $\exists cell_1$, $cell_2$, ..., $cell_r \in
S_k$ such that $|v_1 \cup v_2 \cup...\cup v_{r}|=r$, then $\forall
cell_{ij} \in S_k$ and $cell_{ij} \notin \{cell_1$, $cell_2$, ...,
$cell_r\}$, we have $c_{ij} \in v_{ij}-v_1 \cup v_2 \cup...\cup
v_{r}$.
\end{theorem}

{\em proof:} Vertices $cell_1, cell_2, ..., cell_r, cell_{ij}$ are
in the same section, so $c_1,c_2,...,c_r$ and $c_{ij}$ are all
different,
\begin{equation}\label{Equ:rR_1}
|\{c_1,c_2,...,c_r\}|=r
\end{equation}
\begin{equation}\label{Equ:rR_2}
c_{ij} \in v_{ij} - \{c_1,c_2,...,c_r\}
\end{equation}

According to Definition \ref{Def:filled},
\begin{equation}\label{Equ:rR_3}
\{c_1,c_2,...,c_r\} \subseteq v_1 \cup v_2 \cup...\cup v_r
\end{equation}

So,

\begin{equation}
|\{c_1,c_2,...,c_r\}| \leq |v_1 \cup v_2 \cup...\cup v_r|
\end{equation}

with equation (\ref{Equ:rR_1}) and assumption in the theorem, we
have,
\begin{equation}\label{Equ:rR_4}
r = |\{c_1,c_2,...,c_r\}| \leq |v_1 \cup v_2 \cup...\cup v_r| = r
\end{equation}

it implies that in equation (\ref{Equ:rR_3}),
\begin{equation}\label{Equ:rR_5}
\{c_1,c_2,...,c_r\} = v_1 \cup v_2 \cup...\cup v_r
\end{equation}

substitute for equation (\ref{Equ:rR_2}), we conclude:
\begin{equation}\label{Equ:rR_6}
c_{ij} \in v_{ij}-v_1 \cup v_2 \cup...\cup v_{r}
\end{equation} \hfill $\Box$

This theorem allows player to remove a set of numbers from the
candidates set of a vertex, depending on the candidates sets of
connected vertices in the same section.

Theorem \ref{The:rR} is a very strong reduction rule for the Sudoku
puzzle reasoning. In experimental results, we will show that it can
solve nearly all the Sudoku puzzles, except some very hard puzzles.
Because these very hard puzzles provide less redundant pieces of
information for heuristic reasoning \cite{Henz}. In order to solve
all the puzzles, we have to add the following Inference Rule. This
theorem allows player to make additional assumptions, which can
increase the redundant information for reasoning.

\begin{theorem}[Inference]\label{The:Inference} $\forall
v_{ij}=\{c_1,c_2,...\}$, if we set $v_{ij}=\{c_k\} \subseteq
\{c_1,c_2,...\}$, then if r-Reduction can reduce the puzzle to a
solution, then we conclude $c_{ij}=c_k$.
\end{theorem}
{\em proof:} As we know, $c_{ij} \in v_{ij}$ is a single number. The
puzzle has exactly one solution, so $c_{ij}$ is in the solution.
Because we have set $v_{ij}=\{c_k\}$, $cell_{ij}$ is surely filled
with $c_k$ in the solution. So we conclude $c_{ij} = c_k$. \hfill
$\Box$

Based on Theorem \ref{The:rR} and Theorem \ref{The:Inference}, we
propose the following algorithm that can solve all the Sudoku
puzzles in nearly polynomial time.

\begin{algorithm}[Algorithm for solving Sudoku]
\begin{tabbing}
\\
Init\=iation: set global variables $v_{11},...,v_{99}$, according
to the \\
\>filled cells in Sudoku.\\
\\
bool Solver()\\
\{\\
\>whil\=e(1)\\
\>\{\\
\>\>Reduction();\\
\>\>if \=(Sudoku is not solved)\\
\>\>\>Inference();\\
\>\>else if (Sudoku has no solution)\\
\>\>\>return false;\\
\>\>else return true;\\
\>\}\\
\}\\
\\
Reduction()\\
\{\\
\>r=1;\\
\>while(r $<$ 9)\{\\
\>\>r-Reduction;\\
\>\>if(no candidates set is reduced)\\
\>\>\>r++;\\
\>\>else r=1;\\
\>\>\}\\
\}\\
\\
Inference()\\
\{\\
\>Select randomly a vertex $cell_{ij}$ which was not selected \\
\>\>and has $|v_{ij}|>1$;\\
\>for each $x \in v_{ij}$ \\
\>\{\\
\>\>set $c_{ij}=x$;\\
\>\>if( Solver() ) return;\\
\>\}\\
\}
\end{tabbing}
\end{algorithm}

As we mentioned in Theorem \ref{The:rR}, the ``r-Reduction''
statement in function ``Reduction()'' tries to find $r$ vertices
that share $r$ numbers for reduction, for each section in the 27
sections. Its computational complexity is polynomial.

Notice that if we only use the inference rule, the reasoning process
actually generates a backtracking inference tree, with exponential
complexity. However, if r-Reduction can solve puzzles with few
inferences (e.g., 1 or 2 times), the practical computational
complexity is still nearly polynomial. As we can learn from the
algorithm, we only use the inference as an assistant of r-Reduction.
As we will show in the experiments, the practical computational
complexity is almost equal to the polynomial complexity of
r-Reduction.

\section{On the Game Complexity of Sudoku}
The game complexity of Sudoku depends on the initial distribution of
the numbers in the cells \cite{Lynce}. This section aims at
providing a quantitative metric to assess the complexity without any
playing. In \cite{Shannon}, information entropy is proposed to
evaluate the uncertainty of signals sent by an information source.
The ``uncertainty'' concept is similar to the uncertainty of
choosing the correct number for the vertex when the player is
traversing the Sudoku graph, by considering the possible choices on
a vertex as the possible signals to be sent in a state of
information source. So we propose to evaluate the complexity of
solving Sudoku puzzles using the information entropy.

On the player's viewpoint, there are $|v_{ij}|$ possible choices for
each vertex $cell_{ij}$. At the first step, the player has to choose
one from the $|v_{ij}|$ possible choices. Then, at the second step,
the player chooses to go to one of the 20 connected vertices in the
same row, column and sub-grid (see Fig. \ref{Fig:Sudoku_Graph}).
Taking the two steps together, the player has $20 \times |v_{ij}|$
choices ($(c_{k},e_{ij,lm})$ pairs). The choices are of the same
probability, because the player can only try them one by one without
additional hints. So for each vertex $v_{ij}$, the entropy is
$H_{ij}=-\log \frac{1}{20 |v_{ij}|}$.

For the entire graph, the entropy is the average entropy of all the
81 vertices:

$H=-\frac{1}{81} \sum_{ij} \log \frac{1}{20|v_{ij}|} = - \log
\frac{1}{20} -\frac{1}{81} \sum_{ij} \log \frac{1}{|v_{ij}|}$

The first part ``$- \log \frac{1}{20}$'' of the formula represents
the intrinsic complexity of the puzzle which is associated with the
structure (20 connected vertices). Then the later part of the
formula expresses the complexity related to the distribution of the
initial values. For simplification, we remove the constant part
$-\log \frac{1}{20}$ from the equation, and define the Sudoku
entropy and the game complexity as follows.

\begin{definition}
The \textbf{Sudoku entropy} is defined as
\begin{equation}
H_G=-\frac{1}{81} \sum_{ij} \log \frac{1}{|v_{ij}|}
\end{equation}
\end{definition}

\begin{definition}
The \textbf{game complexity} is the Sudoku Entropy after initiation
plus one 1-reduction.
\end{definition}

The difficulty for players to solve the puzzle is based on the
Sudoku entropy. The game complexity for players is the Sudoku
entropy after the initiation of the graph, and removing the numbers
from vertex candidates set which are obviously impossible choices.

For example, in Fig. \ref{Fig:Sudoku}, after initiation and one
1-reduction, $cell_{31}$ has the candidates set $\{1,2\}$ , so
$H_{31}=-\log \frac{1}{2}$. Another vertex $cell_{13}$ has the
candidates set $\{1,2,4\}$, so $H_{13}=-\log \frac{1}{3}$.

\section{Experimental Results}
In this section, we show the effectiveness of our algorithm for
solving Sudoku and of the quantitative metrics for Sudoku complexity
estimation.\\

Up to now, since difficulty levels are defined by experts, there is
no tool based on an analytical approach for Sudoku estimation. So we
have created test sets of three difficulty levels. The ``easy''
level and the ``intermediate'' level include both 24 instances from
the first 3 books of easy and intermediate Sudoku puzzles
respectively in \cite{Krazydad}.

According to \cite{Lynce}, there are no known puzzles with 16
pre-assigned cells, because they lead to more than one solution. So
the puzzles with only 17 pre-assigned cells are the most difficult
puzzles. When additional cells are assigned, the information
redundancy increases, and the complexity decreases.
\cite{MinimumSudoku} contains 47,621 Sudokus of exact 17
pre-assigned cells. Our ``hard'' level data set includes 10,000
puzzles from them.

Using the reasoning algorithm and the game complexity metric based
on entropy, the results of the 3 levels are compared in Table
\ref{Tab:Result_3l}. The r-Reduction rule is strong enough to solve
all the ``easy'' and ``intermediate'' puzzles, and 70.5\% of the
``hard'' puzzles.

The values of game complexity discriminate the actual difficulty
levels of puzzles perfectly. The easy puzzles have game complexity
values in the domain [0.5362, 0.9046], which are all less than the
values of intermediate puzzles in the domain [1.2246, 1.3965], which
are all less than the values of hard puzzles in the domain [1.6946,
1.8189]. Therefore, our quantitative estimation is highly in
accordant with the evaluation of the experts who design the puzzles.

\begin{table}[hbt]
  \centering
  \begin{tabular}{|c|c|c|c|}
    \hline
    % after \\: \hline or \cline{col1-col2} \cline{col3-col4} ...
                                      &  Easy   & Intermediate & Hard \\
    \hline
    Solved (r-Reduction only)         & 100\%   &   100\%      & 70.5\%\\
    \hline
    Solved (r-Reduction + Inference ) & (100\%) &   (100\%)    & 100\%\\
    \hline
    \hline
    Minimum Game Complexity         &  0.5362 &  1.2246      & 1.6946\\
    \hline
    Maximum Game Complexity         &  0.9046 &  1.3965      & 1.8189\\
    \hline
    Average Game Complexity         &  0.7439 &  1.3049      & 1.7526\\
    \hline
  \end{tabular}
  \caption{Results on 3 levels}\label{Tab:Result_3l}
\end{table}

The detailed results of ``hard'' puzzles are reported in Table
\ref{Tab:Result_hard}. If we only use r-Reduction, 70.5 percent of
the puzzles can be solved. If the inference rule is allowed, all the
puzzles can be solved with only average 1.6 inferences for a puzzle.
So, as explained in section 3, this keeps the practical
computational complexity of the algorithm in nearly polynomial time.
There is nearly no difference on time consumed between the
situations when the inference rule is enabled or not.

The average game complexity of the unsolved puzzles is greater than
that of the solved puzzles, which proves again that our metric can
discriminate the difficulty levels of Sudoku puzzles.\\

\begin{table}[tb]
  \centering
  \begin{tabular}{|c|c|c|c|c|}
    \hline
    % after \\: \hline or \cline{col1-col2} \cline{col3-col4} ...
                & Solved & Unsolved & IT$^*$ & AI$^*$ \\
    \hline
    r-Reduction & 7050(70.5\%) & 2950(29.5\%) & 0 & 0 \\
    only        & GC$^*$ 1.75226 & GC$^*$ 1.75349   &   &   \\
    \hline
    r-Reduction & 10000(100\%) & 0           &  16034 & 1.6034 \\
    + Inference & & & &\\
    \hline
  \end{tabular}\\
  $^*$ IT: Inference Times. AI: Average Inferences. GC: Game Complexity.
  \caption{Results of hard puzzles}\label{Tab:Result_hard}
\end{table}

People are trying to produce more diabolic puzzles to challenge the
reasoning ability of computers. For example, \cite{Henz} reported
that many solvers and algorithms fail to solve the problems from
\cite{Inkala} \cite{Ravel}, such as ``AI Escargot'' puzzle. Using
our algorithm, the computer can solve ``AI Escargot'' with only 91
inferences, which is claimed to be ``the most difficult Sudoku
puzzle known so far''. Moreover, the 21 puzzles from \cite{Ravel}
are solved with 2442 inferences, average 116 inferences for one
puzzle. These results prove that our algorithm can solve all the
known puzzles at any difficulty level with high efficiency.

\section{Conclusion and Future Works}

In this paper, we provided a formal basis for studies and
developments of Sudoku solving systems and complexity evaluation. We
defined the formal graph-based knowledge representation of the
Sudoku puzzle, and proposed a heuristic reasoning algorithm based on
the theorems of r-Reduction and Inference. The r-Reduction theorem
is a very strong heuristic reasoning rule for Sudoku puzzles, and we
have showed that it can solve most of the puzzles itself. In order
to solve the puzzles more complex, we proposed to add the inference
theorem as assistant in the algorithm. Because the r-Reduction rule
is strong enough, few inferences are needed. Even for the hard
puzzles, we only need to infer 1.6 times for each puzzle. This keeps
the practical computational complexity of the algorithm in almost
polynomial time.

It is believed that all the Sudoku puzzles can be solved with only
heuristic reasoning, without backtracking inferences. However, if
only using r-Reduction, 70.5\% of the hard puzzles are solved. So,
for the future works, we are trying to find a stronger rule by
extending r-Reduction, in order to solve all the puzzles using a
single rule.

In fact, we developed an approach based on heuristic reasoning on
certain discrete structure (graph), to solve a constraint
satisfiability problem. This approach may provide a new method and
concrete example for solving constraint-based problems. And it could
also lead to new techniques for developing tutoring systems for
Sudoku learners \cite{Caine2006}.

We also proposed game complexity to quantitatively measure the
difficulty levels of Sudoku puzzles. The concept can be interpreted
very well in the players' viewpoints, and it is also shown to be
effective in experiments. It could also provide a potential
measurement of uncertainty in reasoning.

\end{document}